# A SPEECH CORPUS FOR CHRONIC KIDNEY DISEASE

*Jihyun Mun[1], Sunhee Kim[2], Myeong Ju Kim[3], Jiwon Ryu[4], Sejoong Kim[3,4,5], Minhwa Chung[1]*

[1]Department of Linguistics, Seoul National University, Republic of Korea
[2]Department of French Language Education, Seoul National University, Republic of Korea
[3]Center for Artificial Intelligence in Healthcare, Seoul National University, Republic of Korea
[4]Department of Internal Medicine, Seoul National University Bundang Hospital, Republic of Korea
[5]Department of Internal Medicine, Seoul National University College of Medicine, Republic of Korea

## ABSTRACT

In this study, we present a speech corpus of patients with chronic kidney disease (CKD) that will be used for research on pathological voice analysis, automatic illness identification, and severity prediction. This paper introduces the steps involved in creating this corpus, including the choice of speech-related parameters and speech lists as well as the recording technique. The speakers in this corpus, 289 CKD patients with varying degrees of severity who were categorized based on estimated glomerular filtration rate (eGFR), delivered sustained vowels, sentence, and paragraph stimuli. This study compared and analyzed the voice characteristics of CKD patients with those of the control group; the results revealed differences in voice quality, phoneme-level pronunciation, prosody, glottal source, and aerodynamic parameters.

## 1. INTRODUCTION

A continuous decline in kidney function and structural damage to the kidneys are characteristics of chronic kidney disease (CKD) [1]. CKD is a serious condition with a high prevalence around the world that, if not detected and treated promptly, requires renal replacement therapy, such as dialysis. Although there may not be any symptoms in the early stages of the disease, blood and urine tests are still required to diagnose CKD, so awareness of the condition is still low [1][2]. Therefore, a new index that can both diagnose the disease and gauge its severity should be created. It should also be non-invasive and simple to repeat.

CKD affects a variety of bodily systems, particularly the respiratory system, as well as the cardiovascular, neurological, musculoskeletal, immunological, endocrine, and metabolic systems [3]. The lungs and kidneys both contribute to maintaining the body's acid-base balance in both healthy and diseased states, therefore any changes to the renal system will affect the respiratory system and vice versa [4]. The strength and endurance of the respiratory muscles are significantly reduced in CKD patients compared to non-CKD persons, and the potency of the laryngeal and respiratory muscles is also severely compromised [3][5]. The characteristics of end-stage renal disease (ESRD), which include a buildup of uremic toxins, an acid-base imbalance, and volume overload, are also known to cause a change in voice due to diminished lung function and vocal fold edema [6]. As respiration is the primary source of speech [4], analyzing the voice characteristics of CKD patients and automatically detecting and predicting the severity of CKD through speech may be useful in the early diagnosis and effective treatment of CKD.

Tables 1 and 2 show the parameters, stimuli, and participants of the previous studies, and analysis results of the CKD voice analysis.

**Table 1**. Parameters, stimuli, and participants in previous studies

| Category | | Contents | Papers |
|---|---|---|---|
| Parameters | Voice quality | Jitter | [3, 4, 5, 7] |
| | | Shimmer | [3, 4, 5, 7, 8] |
| | | Harmonics-to-noise ratio (HNR) | [3, 5, 7, 8] |
| | Pitch | Fundamental frequency (F0) | [3, 4, 5, 7, 8] |
| | Aerodynamic | Maximum phonation time (MPT) | [3, 4, 5, 8] |
| Stimuli | | /ipipi/ | [3] |
| | | /a/ | [3, 4, 8] |
| | | /s/, /z/ | [4] |
| Participants | | Non-CKD vs. CKD stage 3-5 (without hemodialysis (HD)) vs. HD | [5] |
| | | Non-CKD vs. HD | [3, 4, 7] |
| | | Non-CKD vs. CKD without HD | [8] |

**Table 2**. CKD voice analysis results

| Category | Parameter | Results |
|---|---|---|
| Voice quality | Jitter | ↑ [3, 4, 7] |
| | | ↓ [5] |
| | Shimmer | ↑ [3, 4, 5, 7, 8] |
| | HNR | ↑ [7, 8] |
| | | ↓ [3, 5] |
| Pitch | F0 | ↑ [3, 4, 5, 7] |
| | | ↓ [8] |
| Aerodynamic | MPT | ↓ [3, 4, 5, 8] |

↑: CKD > non-CKD
↓: CKD < non-CKD

Previous research identified a voice difference between speakers with and without CKD, as demonstrated in the Tables. Finding out the characteristics of the CKD voice was challenging, though, because the results were varied. Additionally, they only looked at a small number of speech-related features using limited voice data. The results, however, can differ since the variables they looked at can be evaluated in sentences rather than only in continuous vowel sounds. Moreover, because CKD can alter various parts of speech, it is necessary to examine speech using a variety of metrics. It is also difficult to understand how CKD affects voice and how voice changes based on the stage of CKD because they did not identify the CKD group according to stage. Furthermore, no research used automatic detection or severity prediction methods for CKD, nor was there a corpus that gathered the voices of CKD patients. Therefore, a speech corpus containing different speech data and participants' information connected to their stage of CKD is needed to understand voice change in CKD patients and construct an appropriate index.

With this goal, this paper introduces a corpus which is developed for studying CKD voice, automatically detecting disease, and predicting severity is introduced. This paper is organized as follows: Section 2 describes the corpus, including participants, metadata, reading script, and recording procedure. Section 3 presents the parameters which are used to analyze the CKD voice, and results & discussion, which are followed by the conclusion in Section 4.

## 2. CORPUS

### 2.1. Participants and metadata

In total, 289 CKD speakers and 14 non-CKD speakers were recruited by us. All of the speakers were chosen from the Bundang Hospital at Seoul National University. The ages of CKD speakers ranged from 23 to 91, with an average of 65 (standard deviation: 14.1), and their severity levels were established according to the doctor's assessment based on eGFR (estimated glomerular filtration rate). The ages of non-CKD speakers ranged from 35 to 85, with an average of 64. (std: 13.7). Table 3 displays the number of speakers by severity and gender. Some speakers have been recorded multiple times (F: 11, M: 25), and the speech data from these speakers will be used in the longitudinal study to examine how a voice changes as the disease progresses. The exclusion criteria included smoking, asthma, and chronic obstructive pulmonary disease, as well as the presence of vocal cord disease and its history. After recruiting the speakers, we collected metadata of the participants. The following information is gathered as meta-data: language disorder presence, gender, birthdate, place of residence, presence and kind of comorbidities, medication usage, physical conditions at the time of recording.

Table 3. Number of speakers

| Group | | Female | Male |
|---|---|---|---|
| CKD | Stage 1 | 20 | 15 |
| | Stage 2 | 38 | 41 |
| | Stage 3 | 46 | 62 |
| | Stage 4 | 12 | 35 |
| | Stage 5 | 5 | 4 |
| | Hemodialysis | 2 | 2 |
| | Transplant | 2 | 5 |
| Non-CKD | | 9 | 4 |

### 2.2. Reading script

First, as in previous studies, participants are required to sustain the vowel /a/. The vowel with the highest first formant, /a/, does not significantly increase the first or second harmonics [9]. Vowel speech can be used to extract voice quality features, pitch features, glottal source parameters, and maximum phonation time (MPT). Second, they are required to read a text made up entirely of vocal sounds. This sentence speech can be used to extract voice quality features and pitch features as in vowel speech because it only contains voiced sounds. We want to examine how these features are represented in sentences. Finally, they were required to read a paragraph made up of six phonetically balanced sentences that varied in length [10]. Spectral features, prosodic features, and phoneme- level pronunciation features are all extracted from the paragraph speech.

### 2.3. Recording procedure

The Seoul National University Bundang Hospital served as the site of the recording. For recording purposes, a Samsung Galaxy S series smartphone and an AKG C414 B-ULS microphone with an AKG PF80 pop filter were both used. The Scarlett Solo Audio Interface was utilized to convert the microphone signals into a computer-readable format. To prevent air puffing, the smartphone and microphone were situated 20 cm from the speaker. A guide led the speaker through the process during each recording session, instructing them to wait for at least three seconds in between each sentence and to re-record any sentences that drastically varied from the prompt. All speakers were asked to speak naturally and to help them do so, they recorded a sample sentence that started with greetings and self-introductions. The speech was recorded as a WAV file with a 16kHz sampling rate. We segmented the utterances into separate WAV files when the recording was done, and Praat did this.

## 3. ANALYSIS RESULTS

### 3.1. Methods

We examined the voices of speakers up to CKD stage 4 because the number of hemodialyses, renal transplantation, and CKD stage 5 speakers was quite low in comparison to speakers at earlier stages. Additionally, there were fewer non-CKD speakers than CKD speakers, thus we created a new classification for speaker severity based on eGFR. Speakers with an eGFR of more than 60 were considered non- CKD speakers (127 participants, 67 females, 60 males), whereas speakers with an eGFR of less than 60 were considered CKD speakers. For CKD speakers, stage 3 was defined as having an eGFR of 30 or more and less than 60 (108 participants, 46 females, 62 males), and stage 4 as having an eGFR of 15 or more and less than 30 (47 subjects, 12 females, 35 males).

The voices of CKD and non-CKD speakers were first compared and examined. On parameters that satisfied the requirements for data normality, the independent sample t-test was used, and the Mann-Whitney U test was used on parameters that did not. Second, the voices of these three groups—those without CKD, those in CKD stage 3, and CKD stage 4—were compared. On parameters that satisfied the requirements for data normality, one-way ANOVA was used, and on parameters that did not, the Kruskal-Wallis H test was used. The Bonferroni post hoc test was additionally conducted. Finally, correlation and regression analyses were conducted to examine the relationship between eGFR and each parameter. All statistical analysis was performed using IBM SPSS Statistics 26 [11].

## 3.2. Speech-related features

Table 4 lists the parameters that were chosen to represent different characteristics of speech in the CKD speech analysis. With regard to the impact of CKD on speech, we utilized [12]'s feature set and added new features, such as aerodynamic and glottal source parameters.

**Table 4.** Speech-related features

| Category | | Features |
|---|---|---|
| Spectral features | | MFCCs |
| Voice quality features | | Jitter, shimmer, HNR, number of voice breaks, degree of voice breaks |
| Prosody features | Pitch | F0 mean/sd/med/min/max |
| | Speech rate | Total duration, speech duration, speaking rate, articulation rate, number of pauses, pause duration |
| | Rhythm | %V, deltas, varcos, rPIVs, nPVIs |
| Phoneme-level pronunciation features | Percentage of correct phonemes | PCC, PCV, PCT |
| | Degree of vowel distortion | VSA, VAI, FCR, F2-ratio |
| Aerodynamic feature | | MPT |
| Glottal source parameters | | H1-H2, H1-A1, H1-A2, H1-A3 |

### 3.2.1. MFCCs

A representation of a sound's short-term power spectrum used in sound processing is called a Mel-frequency cepstrum (MFC), which is based on a linear cosine transform of a log power spectrum on a nonlinear Mel scale of frequency. An MFC is made up of coefficients known as mel-frequency cepstral coefficients (MFCCs). Applications for speaker identification and recognition have usually used MFCCs. Their applicability has been expanded to include speech quality evaluation for medical purposes [13]. Using the librosa [14] toolkit, we extract 12-dim MFCCs and log energy from each speech.

### 3.2.2. Voice quality features

Five voice quality features, jitter, shimmer, harmonic to noise ratio (HNR), number of voice breaks, and degree of voice breaks—were chosen for this investigation. While shimmer, which is very similar to jitter, represents changes in amplitude, jitter represents changes in F0 over time. The HNR is the proportion of harmonic to noise energy. It has been demonstrated that jitter, shimmer, and HNR can be used to describe vocal traits and provide a pathological voice diagnosis [15]. Voice break features reveal the vocal ability to maintain phonation. Using Praat [16], all voice quality features are extracted. The minimum and maximum pitches are respectively set to 70 Hz and 625 Hz for jitter, shimmer, and HNR [12]. The number of voice breaks is the first feature in terms of voice breaking features. Praat divided the pitch floor, which is set at 70 Hz, by the number of intervals between consecutive glottal pulses that are longer than 1.25. The degree of voice breaks is then determined by dividing the sum of voice break duration by speech duration [17].

### 3.2.3. Prosody features

Pitch, speech rate, and rhythm are the three prosody feature categories that are extracted.

Using Praat, we calculate the median, minimum, maximum, mean, and standard deviation of F0 for pitch.

For speech rate, we measure the total duration, speech duration, speaking rate, articulation rate, pause duration, and the number of pauses. Speaking rate is the ratio of syllables generated to total duration, and articulation rate is the ratio of syllables produced to speech duration. We include pause-related variables, such as the number of pauses and pause duration, because CKD decreases respiratory function. Parselmouth [18] is used to extract these features.

We extract %V, deltas, Varcos, rPVIs, and nPVIs for rhythm. The proportion of vocalic utterance intervals is represented as %V. Consonantal and vocalic interval standard deviations are referred to as deltas, and the normalized delta values by the average length of these intervals are called Varcos. The vocalic and consonantal intervals are ordered temporally in the pairwise variability index (PVI). The raw PVI is referred to as rPVI and the normalized PVI as nPVI [19]. Correlatore 2.3.4 [20] is used to extract these features from the data.

### 3.2.4. Phoneme-level pronunciation features

Two categories of phoneme-level pronunciation features are the percentage of correct phonemes and the degree of vowel distortion.

The features of the percentage of correct phonemes include the percentage of correct consonants, the percentage of correct vowels, and the percentage of total correct phonemes (PCT). A speech recognizer that has been trained on speakers without CKD is used to extract these features. The AI Hub corpus [21] is used to train the acoustic model, and the Kaldi toolkit [22] is used for ASR training. The number of matches between a phoneme sequence from an automatic speech recognition model and the canonical pronunciation sequence is used to calculate PCC, PCV, and PCT.

Vowel Space Area (VSA), Vowel Articulatory Index (VAI), Formant Centralized Ratio (FCR), and F2-ratio are indicators of how distorted a vowel is. The region where the first and second formant frequency coordinates (F1, F2) of a vowel are connected by a line in a two-dimensional space is called VSA [23]. The indicators of vowel centralization are VAI and FCR, and they have an antagonistic connection. They have been used to describe changes in vowel articulation as substitute parameters. High FCR and low VAI values are seen when the vowel space is concentrated in relation to the standard coordinates [24][25]. By combining a speech recognizer and Praat with [12]'s methodology, these features are extracted.

### 3.2.5. Aerodynamic feature

The objective measurement of the effectiveness of the respiratory mechanism during phonation is the maximum phonation time (MPT), which is defined as the capacity to maximally sustain a vowel after having taken a maximal inspiration [26]. Praat is used to extract MPT.

### 3.2.6. Glottal source parameters

The term "glottal source" refers to glottal flow, which is air evicted from the lungs and modulated by the vocal folds as it passes down the trachea [27]. We suggest incorporating parameters relating to glottal flow because it is well known that CKD can affect respiration [3][4][5] and that it can result in vocal cord edema [6]. H1-H2, H1-A1, H1-A2, and H1-A3 are the four glottal source parameters that are used [28][29]. The first and second harmonics of the Fourier spectrum are denoted by H1 and H2, respectively. The amplitudes of the first, second, and third formants are denoted by the A1, A2, and A3, respectively. Those parameters are known as acoustic measurements to characterize differences along the glottal constriction continuum [30]. The calculation of VoiceSauce [31], a software that automatically extracts voice measurements from audio recordings, is used to extract glottal source features by Praat.

### 3.3. Results

The statistically significant parameters are displayed in Tables 5, 6, 7, and 8. The parameters measured in the sustained vowel and sentence are referred to as _v and _s, respectively. There were differences in voice quality, phoneme-level pronunciation, prosody, glottal source, and aerodynamic parameters when it was examined whether there was a difference in voice according to the existence and severity of the disease by group comparison. In terms of voice quality, the CKD groups showed lower jitter and shimmer values than the non-CKD group, and the lower the value as the severity of the CKD group increased. Similarly, the CKD groups showed higher values in HNR, and the value increased with increasing severity. According to the data, patients do not distort vowels at the phoneme level, although both vowels and consonants are frequently mispronounced in patients. Patients specifically exhibit greater consonant errors than vowel errors. The CKD groups showed higher pitch in males but lower pitch in females. Additionally, the CKD group showed longer speech duration and, as a result, lower articulation rate.

To understand the impact of eGFR on each parameter, we performed correlation and regression analysis. First, we determine which parameter values rise or fall with eGFR by correlation analysis. There was a significant correlation between eGFR and parameters in the aerodynamic, glottal source, phoneme-level pronunciation, and prosody parameters, similar to the findings of group comparisons. There was a statistically significant positive correlation between eGFR and parameters, except Std F0 v, percent V, and delta-V. Then, using eGFR as an independent variable, we performed a regression analysis to see if eGFR has an impact on the dependent variables. Similar to the findings of the correlation analysis, the findings of the regression analysis demonstrated that eGFR significantly affected the aerodynamic, glottal source, phoneme-level pronunciation, and prosody parameters.

### 3.4. Discussion

Due to the contradictory results of previous studies, as mentioned earlier, it was challenging to identify the characteristics of CKD voice. However, several metrics revealed different results from previous studies. Most previous studies reported larger values in the CKD group for jitter and shimmer, however the experimental results revealed lower values. In terms of fundamental frequency, regardless of gender, [4] reported higher F0 and [8] reported lower F0 in CKD groups. However, the results of the experiment revealed that while the F0 was lower in the CKD group for females, it was higher for males. It suggests that when examining CKD voice, gender should be considered.

**Table 5.** Non-CKD vs. CKD voice analysis

| Category | Parameter | | Non-CKD | CKD | Statistic | p-value |
|---|---|---|---|---|---|---|
| Voice quality | Jitter_s | | 1.83 | 1.72 | Mann-Whitney U=9355.0 | 0.09 |
| | Shimmer_s | | 8.73 | 7.97 | Mann-Whitney U=9131.5 | 0.004 |
| | HNR_s | | 14.58 | 15.39 | t(303)=-2.997 | 0.003 |
| Phoneme-level pronunciation | PCT | | 88.19 | 85.28 | Mann-Whitney U=9078.0 | 0.003 |
| | PCC | | 86.93 | 84.18 | Mann-Whitney U=9342.0 | 0.009 |
| | PCV | | 90.21 | 87.00 | Mann-Whitney U=8809.0 | 0.001 |
| Prosody | Med F0_v | F | 206.72 | 187.38 | Mann-Whitney U=1633.0 | 0.007 |
| | | M | 125.58 | 136.34 | Mann-Whitney U=2475.0 | 0.002 |
| | Mean F0_v | F | 203.51 | 182.03 | Mann-Whitney U=1500.0 | 0.001 |
| | | M | 124.65 | 134.22 | Mann-Whitney U=2474.0 | 0.002 |
| | Std F0_v | F | 21.01 | 29.10 | Mann-Whitney U=1614.0 | 0.041 |
| | Min F0_v | F | 104.45 | 82.57 | Mann-Whitney U=1778.0 | 0.041 |
| | Med F0_s | M | 125.01 | 136.03 | Mann-Whitney U=2302.0 | 0.000 |
| | Mean F0_s | F | 195.58 | 185.52 | t(132)=2.485 | 0.014 |
| | | M | 126.28 | 137.01 | Mann-Whitney U=2374.0 | 0.001 |
| | Min F0_s | M | 70.39 | 75.23 | Mann-Whitney U=2802.0 | 0.041 |
| | Speech duration | | 4.47 | 4.61 | Mann-Whitney U=9575.5 | 0.021 |
| | Articulation rate | | 6.33 | 6.13 | t(303)=2.369 | 0.018 |
| | %V | | 73.40 | 73.92 | t(303)=-2.049 | 0.041 |
| | Delta-V | | 176.30 | 180.73 | Mann-Whitney U=9760.0 | 0.039 |
| Glottal source | H1-A3 | | 32.85 | 30.64 | Mann-Whitney U=9526.0 | 0.018 |

It is interesting to observe that while there were no differences in a sustained vowel between groups, there were differences in the sentence utterance. This indicates the need to investigate CKD patients' voices using a range of utterances. There were differences between the groups even in parameters that had not been examined in previous studies, such as phoneme-level pronunciation, speech speed, rhythm, and glottal source parameters. As a result, employing various speech-related parameters, we should examine different features of CKD speech.

Table 6. Non-CKD vs. CKD stage 3 vs. CKD stage 4

| Category | Parameter | | CKD stage 3 | CKD stage 4 | Statistic | p-value |
|---|---|---|---|---|---|---|
| Voice quality | Jitter_s | | 1.78 | 1.60 | H(2)=11.380 | 0.003 |
| | Shimmer_s | | 8.00 | 7.91 | H(2)=8.477 | 0.014 |
| | HNR_s | | 15.25 | 15.69 | F(2, 302)=5.475 | 0.005 |
| Phoneme-level pronunciation | PCT | | 86.18 | 83.39 | H(2)=13.539 | 0.001 |
| | PCC | | 85.04 | 82.37 | H(2)=10.967 | 0.004 |
| | PCV | | 87.98 | 84.94 | H(2)=15.351 | 0.000 |
| Prosody | Med F0_v | F | 192.15 | 169.58 | H(2)=11.028 | 0.004 |
| | | M | 134.35 | 139.38 | H(2)=10.724 | 0.005 |
| | Mean F0_v | F | 185.70 | 168.33 | F(2, 131)=8.305 | 0.000 |
| | | M | 131.84 | 137.85 | H(2)=11.589 | 0.003 |
| | Std F0_v | F | 28.92 | 29.77 | H(2)=7.733 | 0.021 |
| | Med F0_s | M | 134.02 | 139.09 | H(2)=15.004 | 0.001 |
| | Mean F0_s | F | 186.65 | 181.30 | F(2, 131)=3.238 | 0.042 |
| | | M | 135.76 | 138.91 | H(2)=12.928 | 0.002 |
| Aerodynamic | MPT | | 11.48 | 9.08 | H(2)=12.623 | 0.002 |

## 4. CONCLUSION

In this paper, a speech corpus of CKD patients has been provided. It is a tool for analyzing pathological voice analysis, automatically diagnosing diseases, and estimating disease severity. Totaling 289 CKD speakers and 14 non-CKD speakers, we collected and analyzed their data. The findings revealed that the two groups significantly differed between voice quality, phoneme-level pronunciation, prosody, glottal source, and aerodynamic parameters. Aerodynamic, glottal source, phoneme-level pronunciation, and prosody parameters were significantly correlated with eGFR, and eGFR substantially impacted those parameters.

In this study, only the findings for significant parameters were reported because the purpose of this study is to introduce the corpus we developed and suggest a means to use the corpus. We will examine CKD patients' voices in greater depth in upcoming publications. We'll conduct a classification experiment using a range of deep learning and machine learning models to detect diseases and predict their severity. The support vector machine (SVM) with those statistically significant parameters will be used initially. The SVM classifier is the most used classifier for automatically detecting voice disorders because it works better with small datasets and high-dimensional data [32]. Because the corpus size is small in comparison to other classification tasks, such as image classification, we will explore a variety of deep learning models that excel on small-sized datasets, such as ResNet [33].

## 5. ACKNOWLEDGEMENTS

This research was supported by the MSIT (Ministry of Science and ICT), Korea, under the ITRC (Information Technology Research Center) support program (IITP-2022-2018-0-01833) supervised by the IITP (Institute for Information & Communications Technology Planning & Evaluation).

Table 7. Correlation analysis between eGFR and parameters

| Category | Parameter | Pearson correlation coefficient | p-value |
|---|---|---|---|
| aerodynamic | MPT | .191 | 0.001 |
| Glottal source | H1-H2 | .126 | 0.036 |
| | H1-A1 | .145 | 0.016 |
| | H1-A3 | .153 | 0.011 |
| Phoneme-level pronunciation | PCT | .215 | 0.000 |
| | PCC | .202 | 0.001 |
| | PCV | .232 | 0.000 |
| Prosody | Med F0_v (F) | .285 | 0.002 |
| | Mean F0_v (F) | .289 | 0.002 |
| | Std F0_v (F) | -.194 | 0.036 |
| | Mean F0_s (F) | .231 | 0.012 |
| | Articulation rate | .118 | 0.049 |
| | %V | -.122 | 0.043 |
| | Delta-V | -.136 | 0.024 |

Table 8. Regression analysis between eGFR and parameters

| Category | Parameter | | β | p-value |
|---|---|---|---|---|
| Aerodynamic | MPT | | 0.191 | 0.001 |
| Glottal source | H1-H2 | | 0.126 | 0.036 |
| | H1-A1 | | 0.145 | 0.016 |
| | H1-A3 | | 0.153 | 0.011 |
| Phoneme-level pronunciation | PCT | | 0.215 | 0.000 |
| | PCC | | 0.202 | 0.001 |
| | PCV | | 0.232 | 0.000 |
| Prosody | Med F0_v | F | 0.285 | 0.002 |
| | | M | -0.217 | 0.006 |
| | Mean F0_v | F | 0.289 | 0.002 |
| | | M | -0.202 | 0.010 |
| | Std F0_v | F | -0.194 | 0.036 |
| | Max F0_v | M | 0.166 | 0.036 |
| | Med F0_s | M | -0.275 | 0.000 |
| | Mean F0_s | F | 0.231 | 0.012- |
| | | M | -0.233 | 0.003 |
| | Min F0_s | M | -0.185 | 0.020 |
| | Articulation rate | | 0.118 | 0.049 |
| | %V | | -0.122 | 0.043 |
| | Delta-V | | -0.136 | 0.024 |

# 6. REFERENCES

[1] Webster, A. C., Nagler, E. V., Morton, R. L., and Masson, P., "Chronic kidney disease," *The lancet*, vol. 389, no. 10075, pp. 1238-1252, Nov. 2016.

[2] Kwon, S. H., and Han, D. C., "Diagnosis and screening of chronic kidney disease," *Korean Journal of Medicine*, vol. 76, no. 5, pp. 515-520, 2009.

[3] Hassan, E. S., "Effect of chronic renal failure on voice: an acoustic and aerodynamic analysis," *The Egyptial Journal of Otolaryngology*, vol. 30, no. 1, pp. 53-57, Aug. 2013.

[4] Radish B. Kumar and Jayashree S. Bhat, "Voice in Chronic Renal Failure," *Journal of Voice*, vol. 24, no. 6, pp. 690-693, Nov. 2010.

[5] Abd El-gaber, F. M., Sallan, Y., and El Sayed, H. M. E., "Acoustic Characteristics of Voice in Patients with Chronic Kidney Disease," *International Journal of General Medicine*, vol. 14, pp. 2465-2473, June 2021.

[6] Jung, S. Y., Ryu, J. H., Park, H. S., Chung, S. M., Ryu, D. R., and Kim, H. S., "Voice change in end-stage renal disease patients after hemodialysis: correlation of subjective hoarseness and objective acoustic parameters," *Journal of Voice*, vol. 28, no. 2, pp. 226-230, July 2013.

[7] Zaky, E. A., Mamdouh, H., Esmat, O., and Khalaf, Z., "Voice problem in a patient with chronic renal failure," *The Egyptian Journal of Otolaryngology*, vol. 36, no. 1, pp. 1-8, Nov. 2020.

[8] Mudawwar, W. A., Alan, E. S., Sarieddine, D. S., Turfe, Z. A., and Hamdan, A. L. H., "Effect of renal failure on voice," *ENT: Ear, Nose & Throat Journal*, vol. 96, no. 1, pp. 32-36, Jan. 2017.

[9] Ahn, H. K. (2000). The H1*-H2* Measure. *Speech Sciences*, 7(2), 85-95.

[10] Lee, S. E., Kim, H., Sim, H. S., Nam, C. M., Choi, J. Y., and Park, E. S., "Auditory-perceptual evaluation of the speech of adults with hearing impairment based on suprasegmental factors, speech intelligibility, and speech acceptability", *Communication Sciences & Disorders*, vol. 15, no. 4, pp. 477-493, Nov. 2010.

[11] IBM Corp. Released 2019. IBM SPSS Statistics for Windows, Version 26.0. Armonk, NY: IBM Corp

[12] Yeo, E. J., Kin, S., and Chung, M., "Automatic Severity Classification of Korean Dysarthric Speech Using Phoneme-Level Pronunciation Features," *in Proc. Interspeech 2021*, 2021, pp. 4838-4842, doi: 10.21437/Interspeech.2021-1353.

[13] Benba, A., Jilbab, A., Hammouch, A., and Sandabad, S., "Voiceprints analysis using MFCC and SVM for detecting patients with Parkinson's disease," in *2015 International conference on electrical and information technologies (ICEIT)*, 2015, pp. 300-304.

[14] B. McFee, C. Raffel, D. Liang, D. P. Ellis, M. McVicar, E. Battenberg, and O. Nieto, "librosa: Audio and music signal analysis in python," in *Proceedings of the 14th python in science conference*, vol. 8, pp. 18-25, July 2015.

[15] Teixeira, J. P., Oliveira, C., and Lopes, C., "Vocal acoustic analysis-jitter, shimmer and hnr parameters," *Procedia Technology*, vol. 9, pp. 1112-1122, Dec. 2013.

[16] P. Boersma, "Praat, a system for doing phonetics by computer," *Glot International*, vol. 5, no. 9/10, pp. 341-345, 2001.

[17] Hernandez, A., Yeo, E. J., Kim, S., & Chung, M. (2020, September). Dysarthria Detection and Severity Assessment Using Rhythm-Based Metrics. In *INTERSPEECH* (pp. 2897-2901).

[18] Y., Jadoul, B. Thompson, & B. de Boer, "Introducing Parselmouth: A Python interface to Praat," *Journal of Phonetics*, 71, 1-15, 2018. https://doi.org/10.1016/j.wocn.2018.07.001

[19] Hernandez, A., Kim, S., & Chung, M. (2020). Prosody-based measures for automatic severity assessment of dysarthric speech. *Applied Sciences*, 10(19), 6999.

[20] P. Mairano, and A. Romano, "Un confronto tra diverse metriche ritmiche usando Correlatore," In: Schmid, S., Schwarzenbach, M. & Studer, D. (eds.) La dimensione temporale del parlato, (Proc. of the V National AISV Congress, University of Zurich, Collegiengebaude, 4-6 February 2009), Torriana, pp. 79-100, 2010.

[21] AIHub Homepage. Available online: http://www.aihub.or.kr/aidata/105 (accessed on 5 August 2020).

[22] D. Povey, A. Ghoshal, G. Boulianne, L. Burget, O. Glembek, N. Goel, and K. Vesely, "The Kaldi speech recognition toolkit," In *IEEE 2011 workshop on automatic speech recognition and understanding*, 2011.

[23] Sandoval, S., Berisha, V., Utianski, R. L., Liss, J. M., and Spanias, A., "Automatic assessment of vowel space area," *The Journal of the Acoustical Society of America*, vol. 134, no. 5, pp. EL477-EL483, Nov. 2013.

[24] Skodda, S., Visser, W., and Schlegel, U., "Vowel articulation in Parkinson's disease," *Journal of voice*, vol. 25, no. 4, pp. 467-472, July 2011.

[25] Kim, S., Kim, J. H., Ko, D. H., Kim, S., Kim, J. H., and Ko, D. H., "Characteristics of vowel space and speech intelligibility in patients with spastic dysarthria," *Communication Sciences & Disorders*, vol. 19, no. 3, pp. 352- 360, Sep. 2014.

[26] Speyer, R., Bogaardt, H. C., Passos, V. L., Roodenburg, N. P., Zumach, A., Heijnen, M. A., Baihens, L. W., Fleskens, S. J., and Brunings, J. W., "Maximum phonation time: variability and reliability," *Journal of Voice*, vol. 24, no. 3, pp. 281-284, Oct. 2008.

[27] Drugman, T., Dubuisson, T., and Dutoit, T., "On the mutual information between source and filter contributions for voice pathology detection," 2001. [Online]. Available: arXiv:2001.00583.

[28] Kumar, B. R., Bhat, J. S., and Mukhi, P., "Vowel harmonic amplitude differences in persons with vocal nodules," *Journal of Voice*, vol. 25, no. 5, pp. 559-561, 2011.

[29] Lee, S. J., Cho, Y., Song, J. Y., Lee, D., Kim, Y., and Kim, H., "Aging effect on Korean female voice: Acoustic and perceptual examinations of breathiness," *Folia Phoniatrica et Logopaedica*, vol. 67, no. 6, pp. 300-307, 2015.

[30] Keating, P. A., and Esposito, C., "Linguistic Voice Quality," UCLA Working Papers in Phonetics, 2006. [Online]. Available: https://linguistics.ucla.edu/people/keating/Keating_SST2006_paper.pdf

[31] Shue, Yen-Liang, "The Voice Source in Speech Production: Data, Analysis and Models," Ph.D. dissertation, Dept. Elect. Eng., Univ. California, Los Angeles., 2010.

[32] Hedge, S., Shetty, S., Rai, S., and Dodderi, T., "A survey on machine learning approaches for automatic detection of voice disorders," *Journal of Voice*, vol. 33, no. 6, pp. 947.e11-947.e33, July 2018.

[33] He, K., Zhang, X., Ren, S., & Sun, J. (2016). Deep residual learning for image recognition. In *Proceedings of the IEEE conference on computer vision and pattern recognition* (pp. 770-778).